\documentclass[pdflatex,sn-mathphys-num]{sn-jnl}

\usepackage[namelimits]{amsmath}
\usepackage{amssymb, pifont}
\usepackage{multirow}
\usepackage{tabularx}
\usepackage{makecell}
\usepackage{rotating}
\usepackage{fontawesome}
\usepackage{subfigure}
\usepackage{setspace}
\usepackage{array}
\usepackage[table,xcdraw]{xcolor}
\usepackage[normalem]{ulem}
\useunder{\uline}{\ul}{}
\usepackage{float}
\usepackage[algoruled,lined,boxed,commentsnumbered]{algorithm2e}
\usepackage{enumitem}
\usepackage{caption}
\usepackage{dsfont}
\usepackage{booktabs} 
\usepackage[switch]{lineno}
\usepackage{diagbox}
\usepackage{url}
\usepackage{booktabs}    
\usepackage{xcolor}      
\usepackage{colortbl}    
\usepackage{threeparttable}
\usepackage{arydshln}   
\usepackage{subcaption} 
\usepackage{mathtools}

\theoremstyle{thmstyleone}%
\theoremstyle{thmstyletwo}%

\theoremstyle{thmstylethree}%
\begin{document}

\title[Article Title]{Significance and Stability Analysis of Genotype-Environment Interaction using GxEStat}


\author[1,2]{\fnm{Meng'en} \sur{Qin}}\email{mengenching@gmail.com}

\author[2]{\fnm{Zhe} \sur{Li}}

\author[4]{\fnm{Hui} \sur{Huang}}

\author*[3]{\fnm{Xihong} \sur{Liu}}\email{liuxihong\_ocgk@szgmrmyy.cn}

\affil[1]{\orgdiv{School of Mathematics and Statistics}, \orgname{Henan University}, \orgaddress{\city{Kaifeng}, \country{China}}}

\affil[2]{\orgdiv{Faculty of Computer Science and Artificial Intelligence}, \orgname{Shenzhen University of Advanced Technology}, \orgaddress{\city{Shenzhen}, \country{China}}}

\affil[3]{\orgname{Shenzhen University of Advanced Technology General Hospital}, \orgaddress{\city{Shenzhen}, \country{China}}}

\affil[4]{\orgdiv{School of Nursing}, \orgname{Shanghai Jiao Tong University}, \orgaddress{\city{Shanghai}, \country{China}}}



\abstract{
Genotype-environment (G×E) interactions can influence the performance of genotypes across diverse environments, limiting the reliability of genotype evaluation and selection in breeding programs. In-depth analysis of G×E interactions is therefore essential for understanding how genetic advantages or defects are expressed under varying environmental conditions and for identifying superior and stable genotypes. This study presents an integrated computational framework for G×E analysis that combines significance and stability evaluation within a unified analytical workflow. The framework incorporates the mixed effect modeling to quantify the significance of environment, genotype, and G×E effects, together with multiple stability analysis approaches for characterizing genotype adaptability, environmental representativeness, and performance consistency across environments. 
To support reproducible statistical analysis, we develop GxEStat, an interactive R platform that automates model construction, parameter estimation, statistical inference, and graphical visualization. GxEStat substantially improves analytical efficiency, reproducibility, and accessibility for multi-environment trial analysis. Applications to real breeding datasets demonstrate the effectiveness of the proposed framework for practical breeding research. Codes are available at \url{https://github.com/mason-ching/GxEStat}.
}




\maketitle
\section{Introduction}
G×E interactions refer to the modulation of genetic effects by environmental conditions, which can alter genotype performance across environments and reduce the predictability of phenotypes in a target setting~\cite{watermelon_breeding}. A thorough understanding of organismal behavior requires not only genetic information but also knowledge of the environments in which organisms develop and persist. In-depth investigation of G×E interactions helps clarify how phenotypic traits are shaped by the interplay between genetic background and environmental context, and how genetic advantages or limitations are expressed or masked under specific conditions. Such analyses support gene selection, animal and plant breeding, pharmacogenomics, conservation biology, and related fields.

Biostatisticians and breeders often use analysis of variance (ANOVA) to assess the existence and magnitude of genetic effects and G×E interactions. When analyzing a group of elite varieties, researchers often treat genotypes as fixed terms and environments as random. However, if the objective is to estimate breeding values or yield using the best linear unbiased prediction (BLUP), genotypes are more appropriately treated as random effects and environments as fixed~\cite{gxe_sas,rgxe}. Some statisticians also model genotypes as random effects whenever the goal is to identify the best-performing varieties~\cite{mixed_model}. ANOVA quantifies variance components attributable to fixed and random effects, such as location, genotype, year, replication, and their interactions. Nevertheless, ANOVA has drawbacks in characterizing genotype responses to environmental variation, including the assumption of homogeneous variance across environments~\cite{statistical_analysis_yield_trial}. When the genetic effect and the G×E interaction are both significant, additional stability statistics then need be computed to further evaluate genotype performance.

So far, researchers in related fields have put forward several statistical methods for stability analysis. Regression statistics as the stability measurement were first proposed by Yates et al.~\cite{analysis_group_exp} and later refined by Eberhart et al.~\cite{stability_parameter}. In this method, stability is considered as the mean trait values, the slope of the regression line, and the regression deviation. A high trait mean is a prerequisite for desirable stability. The slope reflects the response of a specific genotype to the environmental factor, which is typically deduced from the mean performance of all genotypes in each environment. If the slope is not significantly different from 1, the genotype is considered broadly adaptable across environments. A slope greater than 1 indicates greater sensitivity to environmental change, corresponding to lower stability but better adaptation to high-yield environments. A slope less than 1 indicates greater resistance to environmental fluctuation, corresponding to higher stability and stronger adaptation to low-yield environments. Perkins and Jinks proposed a related regression coefficient by regressing environmental effects on the interaction between genotype and environment~\cite{environmental_component}.

Several variance-based statistics have also been introduced to measure genotype stability. Wricke designed the stability ecovalence ($W_i^2$) in 1962~\cite{evaluation_method_field_trial}, which represents a genotype’s contribution to the sum of squares of G×E interactions across all environments. However, because $W_i^2$ is defined as a sum of squares, it cannot be directly subjected to a significance test. Shukla proposed the stability variance ($\sigma_i^2$), an unbiased estimate of the variance associated with genotype-environment interactions plus the error linked to genotypes~\cite{stability_analysis_potato}. $\sigma_i^2$ is a linear transformation of Wricke’s ecovalence ($W_i^2$). Genotypes with smaller $\sigma_i^2$ are therefore regarded as more stable. Kang et al. reported that the rank correlation between $W_i^2$ and $\sigma_i^2$ is 1, implying that the two statistics are equivalent for ranking genotype stability. Kang further developed yield stability ($YS_i$) in 1993~\cite{simultaneous_selection}. This nonparametric index combines trait mean and Shukla’s stability variance. It assigns equal weight to mean performance and $\sigma_i^2$, and a genotype is considered stable if its $YS_i$ exceeds the average $YS_i$~\cite{yield_stability_bean,yield_stability_maize}. In this study, $YS_i$ is computed according to the procedure described by Mekbib~\cite{yield_stability_bean}. The stability concept $P_i$ proposed by Lin et al. in 1988 is based on the average sum of squares of years nested within locations~\cite{method_cly}. Smaller $P_i$ values indicate greater stability, i.e., lower temporal variation in trait values. Francis et al. utilized the coefficient of variation ($CV_i$) in 1978~\cite{yield_stability_short_maize}; genotypes with smaller coefficients of variation are considered more stable.

For multi-factor stability analysis, the additive main effects and multiplicative interaction model (AMMI) and genotype main effect plus genotype-environment interaction model (GGE) have become increasingly popular because of their strong graphical interpretability~\cite{multienvironment_trial}. Although both methods often yield similar conclusions, researchers continue to debate the most appropriate way to analyze multi-environment breeding data~\cite{statistical_analysis_ammi_gge,gge_biplot_ammi}. Yan et al. invented the GGE biplot based on the singular value decomposition of an environment-centered two-way data matrix of genotypes and environments~\cite{cultivar_evaluation}. We typically construct the GGE biplot by using the first two principal components (PC1 and PC2), which capture the largest part of the variation in the data. It provides a visual representation of the two-way data matrix and reveals the relationships and interactions between genotype and environment~\cite{gge_biplot}. In the GGE framework, genotype effect and G×E interactions are the two major sources of variation relevant to genotype evaluation and environment characterization. The AMMI integrates ANOVA with principal component analysis: ANOVA is used to model the main effects of gene and environment, whereas PCA is applied to the interaction component, yielding interaction principal components (IPCs)~\cite{statistical_analysis_ammi,ammi_two_maize}. The AMMI model can separate genotypes according to the IPC scores, making it convenient to assess genotype stability and environmental adaptability qualitatively. The closer a genotype’s IPC score is to $0$, the more stable it is in the test environment. Although many statistics have been proposed to evaluate stability from different perspectives, no single method can fully characterize genotype responses across environments. In practice, stability measures are most informative when considered together with trait performance, such as average yield.

Regarding the implementation, Piepho released a SAS program in 1999 for computing single-factor stability statistics~\cite{stability_sas}. Hussein et al. also provided SASGxESTAB in 2000 for calculating single-factor stability statistics and their correlations~\cite{sasgestab_gxe}. However, these SAS programs are no longer readily available due to code problems and version incompatibilities. More recently, Dia et al. presented new SAS codes for G×E interaction analysis~\cite{gxe_sas}. But all of these implementations remain largely programming-oriented and require substantial coding expertise, making them less accessible to breeders and agronomists. Furthermore, existing implementations are typically distributed across different analytical environments, resulting in fragmented workflows, inconsistent analytical procedures, and limited reproducibility.
To address these limitations, we present an integrated statistical computing framework that unifies significance and stability analysis within a standardized analytical workflow for G $\times$ E interaction studies. The framework provides a coherent pipeline for statistical modeling, parameter estimation, inference, visualization, and interpretation, enabling comprehensive G×E analysis in a reproducible and user-friendly manner. We further implement the proposed framework in GxEStat, an open-source interactive R platform that substantially simplifies multi-environment trial analysis while improving analytical efficiency and reproducibility.
The main contributions of this work are as follows:
\begin{itemize}
\item 
First, we propose a thorough G×E analytical framework that unifies significance testing, variance component estimation, and stability assessment, providing a coherent statistical workflow for multi-environment breeding trials.

\item 
Second, we extend classical stability analysis by combining single-genotype group regression with multi-genotype matrix-factorization approaches, allowing genotype performance to be characterized from individualized and population-level perspectives.

\item 
Third, we implement the proposed framework as GxEStat, an interactive and reproducible R package that consolidates analysis, visualization, and interpretation into one platform, substantially lowering the analysis barrier for breeders and agronomists while improving analytical efficiency and usability.
\end{itemize}

\section{Methods}
\subsection{Preliminaries}
\subsubsection{Fixed and Random Effects}
In mixed effect models, fixed and random effects are two key elements used to describe different parameter types in the regression model. First, consider the traditional linear regression equation:
\begin{equation}
\label{equ:linear_model}
    y_{it} = \alpha + \beta x_{it} + \varepsilon_{i},
\end{equation}
where \(t\) denotes $t$-th period, \(i\) represents $i$-th individual, \(\alpha\) and \(\beta\) are the regular intercept and slope. Consider a specific case where \(y_{it}\) denotes the yield of the \(i\)-th variety in the \(t\)-th year and \(x_{it}\) is the scale of the \(i\)-th variety in the \(t\)-th year, and \(\varepsilon_{i}\) is the random error term. Equation (\ref{equ:linear_model}) shows that the yield of each variety is determined by a constant (\(\alpha\)), which will be adjusted according to the scale of the variety (\(\beta x_{it}\)). It seems reasonable to use the general linear regression equation to model this example with a fixed effect of the variety scale. However, the fact that the variety scale (\(x_{it}\)) changes within each variety is ignored, which will lead to the independence of the model residuals and violate the basic assumption of regression.

In order to ensure the independence of the residuals and considering the fact that each variety has multiple observations, it is necessary to separate and estimate the individual characteristic quantity of each variety from the residual (\(\varepsilon_{i}\)). Thus, the following regression equation is obtained:
\begin{equation}
    y_{it} = \alpha + \beta x_{it} + \gamma_{i} + \varepsilon_{it}.
\end{equation}
This equation takes into account the random effects of individual characteristics \(\gamma_{i}\), allowing each variety to have a separate intercept term (\(\alpha + \gamma_{i}\)). Therefore, this model is called the random intercept and fixed slope regression model. If analyzed more deeply, the individual random effects can be split into the random intercept component and a random slope part:
\begin{equation}
    y_{t} = \alpha + \beta x_{it} + \alpha_{i} + \beta_{i}x_{it} + \varepsilon_{it} = \left( \alpha + \alpha_{i} \right) + \left( \beta + \beta_{i} \right)x_{it} + \varepsilon_{it}.
\end{equation}
In this equation, each variety has a separate intercept (\(\alpha + \alpha_{i}\)) and slope (\(\beta + \beta_{i}\)); hence, it is called the random intercept and random slope model. At this point, each individual is estimated to have a unique trajectory. The intercept of these trajectories consists of the group horizontal intercept plus the deviation of the individual random effect in the intercept term, and the slope of these trajectories consists of the group horizontal slope plus the deviation of the individual random effect in the slope term. Note that the individual random effect may be positive or negative because it indicates the way a given individual deviates from the group.

In summary, fixed effects are the group-level intercept and slope (\(\alpha + \beta x_{it}\)). These effects are traditional main effects and interactions, and they are fixed factors that do not change randomly. Random effects are individual characteristic parts (\(\alpha_{i} + \beta_{i}x_{it}\)), which explain the random variability across samples and are randomly selected factors in a certain range in experiments or research. In Section \ref{sec:significance_analysis}, when analyzing the variability of gene-environment interaction, different factors affecting phenotypic response values are regarded as fixed effects or random effects depending on the research purposes.

\subsubsection{PCA and SVD}
Principal component analysis (PCA) is the main technique for dimensionality reduction and multivariate statistical analysis. When a dataset contains a large number of correlated quantitative variables, direct visualization and interpretation become increasingly difficult because each variable represents a separate dimension. PCA addresses this problem by mapping the original data into a new set of orthogonal "principal components" (PCs), which are linear combinations of the original variables. The PCs are arranged according to the amount of variance they explain, allowing original high-dimensional data to be represented by only the first few components while retaining most of the information.
Geometrically, PCA seeks a sequence of mutually orthogonal directions that successively maximize the variance of the projected data. The first PC corresponds to the direction of maximum variance in the original data; the second PC is orthogonal to the first and explains the largest remaining variance, and subsequent principal components are determined analogously. Since the leading PCs generally capture the majority of the total variability, the remaining components often contribute little additional information and may therefore be discarded with minimal information loss.

Traditionally, PCA can be formulated through the eigendecomposition of the covariance matrix of the centered data. Eigenvectors define the PC directions, whereas the corresponding eigenvalues denote the variances explained by each PC. In practice, however, modern numerical implementations usually compute PCA by employing singular value decomposition (SVD) directly to the centered data matrix because SVD is computationally more stable and efficient, particularly for large or non-square matrices.
For any matrix $A\in\mathbb{R}^{m\times n}$, its singular value decomposition is given by
\begin{equation} 
A = U\Sigma V^{T}, 
\end{equation}
where $V\in\mathbb{R}^{n\times n}$ and $U\in\mathbb{R}^{m\times m}$ are orthogonal matrices containing the right and left singular vectors, and $\Sigma\in\mathbb{R}^{m\times n}$ is a diagonal matrix whose diagonal elements are the singular values ordered in descending sequence. Singular values are the square roots of the nonzero eigenvalues of both $AA^{T}$ and $A^{T}A$. Consequently, the columns of $V$ associated with the largest singular values correspond to the PC directions.

Dimensionality reduction is achieved by retaining only the first $k$ singular values and their corresponding right singular vectors, thereby constructing a rank-$k$ approximation of the original data. The transformed low-dimensional representation is obtained by projecting the original data onto the subspace spanned by these leading PCs. Although SVD simultaneously provides low-rank approximations in both the row and column spaces of the data matrix, PCA primarily exploits the right singular vectors to identify the principal component directions used for dimensionality reduction.
\begin{table}[t]
    \setlength{\tabcolsep}{3pt}
    \centering
    \caption{Degree of freedom and effect category of variation source terms.}
    \label{tab:effect_category}
    \small
    \begin{tabular}{lcccccccc}
        \toprule
        \multirow{2}{*}{Source of variation}
        & \multirow{2}{*}{Degree of freedom}
        & \multicolumn{6}{c}{F or R} \\
        \cmidrule(lr){3-8}
        & &\textbf{M1} & \textbf{M2} & \textbf{M3} & \textbf{M4} & \textbf{M5} & \textbf{M6}\\
        \midrule
        genotype & $G-1$ & R & F & F & R & R & F \\
        location & $L-1$ & R & F & R & F & R & F \\    
        year & $Y-1$ & R & F & R & R & F & R \\       
        replication (year$\times$location) & $(R-1) \times L \times Y$ & R & R & R & R & R & R \\   
        location  $\times$ year & $(L-1) \times (Y-1)$ & R & F & R & R & R & R \\
        genotype $\times$ location & $(G-1) \times (L-1)$ & R & F & R & R & R & F \\ 
        genotype $\times$ year & $(G-1) \times (Y-1)$ & R & F & R & R & R & R \\
        genotype $\times$ year $\times$ location & $(G-1) \times (L-1) \times (Y-1)$ & R & F & R & R & R & R \\
        \bottomrule
    \end{tabular}
    \vspace{2mm}
    \footnotesize
    \textbf{Note:} M, model; R, random effect; F, fixed effect.
\end{table}
\subsection{Significance Analysis of Genotype-Environment Interaction}
\label{sec:significance_analysis}
\subsubsection{Mixed Effect Model}
Generally, \(G\) varieties growing in \(E\) environments with \(R\) random replications generate \(G \times E \times R\) observations. The "env" is usually a combination of location and year. Each combination of environment and genotype is called a "treatment", so there is a total of \(G \times E\) treatments.

The mixed effect model should include four factors: location, year, genotype, and replication. According to different purposes, genotype, location, and year can be defined as random or fixed effects, as shown in Table \ref{tab:effect_category}, where L and Y are the numbers of locations and years. 
The specification of fixed and random effects depends on the inferential objective. Genotypes are treated as fixed effects when the objective is to compare specific cultivars, whereas they are modeled as random effects when estimating variance components, breeding values, or genetic gains. Similarly, environments may be considered fixed when specific locations are of direct interest and random when they are viewed as a representative sample of a broader target population. Years and replicate trials are usually regarded as random factors.

We can deduce different equations for different mixed effect models, as shown in Table \ref{tab:mixed_model_spec}, where \(*\) shows the interaction of different terms, i.e., the combination of grouping variables. We note that slopes are only meaningful for continuous independent variables, because a slope measures the rate of change in the response per unit change in the independent variables. For categorical variables, there is no slope – only an intercept shift for each level, which can be modeled as a fixed or random intercept.

We use the \texttt{lmer} function in the \texttt{lme4} package \cite{lme4} to fit linear mixed effect models. The fitted model provides estimates and standard errors for the fixed effect coefficients, together with estimates of the variance associated with the random terms. To obtain predictions for a given trait, the random effects are estimated using the best linear unbiased prediction (BLUP). Under the linear mixed effect model, the BLUP of the random effects is computed by the conditional expectation
\begin{equation}
\widehat{u}=E(u\mid y)=GZ^{T}V^{-1}(y-X\widehat{\beta}),
\end{equation}
where
\begin{equation}
V=ZGZ^{T}+R.
\end{equation}
Here, $G$ denotes the covariance matrix of the random effects, $Z$ is the random-effect design matrix, $V$ is the marginal covariance matrix of the response, $R$ is the residual covariance matrix, $X$ is the fixed-effect design matrix, and $\widehat{\beta}$ is the estimated vector of fixed effect coefficients.

In the proposed package, \texttt{ranef} function of the \texttt{lme4} package \cite{lme4} is applied to extract the BLUPs (conditional means) of the random effects. Owing to the shrinkage property of mixed-effect models, these predictions are typically pulled toward the overall population mean, thereby improving estimation accuracy, particularly for groups with limited observations. For each genotype, the corresponding conditional fitted value is obtained by combining the predicted fixed effects with the estimated random effects, providing the genotype-specific prediction.
\begin{table}[htbp]
    \centering
    \caption{Mixed effect model specifications.}
    \label{tab:mixed_model_spec}
    \small
    \begin{tabular}{c p{0.75\linewidth}}
        \toprule
        \textbf{Model} & \textbf{Mixed-effect model} \\
        \midrule
        Model 1 &
        $\begin{aligned}[t]
        \label{alg:model_1}
        \texttt{Trait} \sim\;& 1
         +(1|\texttt{year})
         +(1|\texttt{location})
         +(1|\texttt{genotype}) \\
        & +(1|\texttt{year:location})
         +(1|\texttt{year:genotype}) \\
        & +(1|\texttt{location:genotype}) \\
        & +(1|\texttt{year:location:genotype}) \\
        & +(1|\texttt{year:location:replication})
        \end{aligned}$
        \\[1em]
        \addlinespace[0.5em] 

        Model 2 &
        $\begin{aligned}[t]
        \texttt{Trait} \sim\;&
        \texttt{year}
        *\texttt{location}
        *\texttt{genotype} \\
        & +(1|\texttt{year:location:replication})
        \end{aligned}$
        \\[1em]
        \addlinespace[0.5em] 

        Model 3 &
        $\begin{aligned}[t]
        \texttt{Trait} \sim\;&
        1
         +(1|\texttt{year})
         +(1|\texttt{location})
         +\texttt{genotype} \\
        & +(1|\texttt{year:location})
         +(1|\texttt{year:genotype}) \\
        & +(1|\texttt{location:genotype}) \\
        & +(1|\texttt{year:location:genotype}) \\
        & +(1|\texttt{year:location:replication})
        \end{aligned}$
        \\[1em]
        \addlinespace[0.5em] 

        Model 4 &
        $\begin{aligned}[t]
        \texttt{Trait} \sim\;&
        1
         +(1|\texttt{year})
         +\texttt{location}
         +(1|\texttt{genotype}) \\
        & +(1|\texttt{year:location})
         +(1|\texttt{year:genotype}) \\
        & +(1|\texttt{location:genotype}) \\
        & +(1|\texttt{year:location:genotype}) \\
        & +(1|\texttt{year:location:replication})
        \end{aligned}$
        \\[1em]
        \addlinespace[0.5em] 

        Model 5 &
        $\begin{aligned}[t]
        \texttt{Trait} \sim\;&
        1
         +\texttt{year}
         +(1|\texttt{location})
         +(1|\texttt{genotype}) \\
        & +(1|\texttt{year:location})
         +(1|\texttt{year:genotype}) \\
        & +(1|\texttt{location:genotype}) \\
        & +(1|\texttt{year:location:genotype}) \\
        & +(1|\texttt{year:location:replication})
        \end{aligned}$
        \\[1em]
        \addlinespace[0.5em] 

        Model 6 &
        $\begin{aligned}[t]
        \texttt{Trait} \sim\;&
        1
         +(1|\texttt{year})
         +\texttt{location}
         +\texttt{genotype} \\
        & +(1|\texttt{year:location}) 
          +(1|\texttt{year:genotype}) \\
        & +\texttt{location:genotype} \\
        & +(1|\texttt{year:location:genotype}) \\
        & +(1|\texttt{year:location:replication})
        \end{aligned}$
        \\
        \bottomrule
    \end{tabular}
\end{table}
\subsubsection{Significance Analysis}
After fitting the linear mixed effect model, the significance of each model component should be tested to identify the sources of phenotypic variation. This step facilitates model simplification by eliminating statistically insignificant effects and provides the necessary basis for the subsequent genotype stability analysis. In addition, the estimated variance components quantify the relative contributions of genotypes, environments, and their interactions to the overall phenotypic variation.

Statistical inference for the fixed effects is performed using the F-test. Although the Satterthwaite approximation is commonly employed to estimate the denominator degrees of freedom for linear mixed models, its performance may deteriorate for models with complex random effects or higher-order interaction terms. Therefore, the \texttt{mixed} function in the \texttt{afex} package \cite{afex} is adopted to perform Type III F-tests based on the Kenward--Roger approximation, which generally provides more accurate finite-sample computation for linear mixed models.

Inference for random effects is more challenging because variance components are constrained to be non-negative. Consequently, the significance of each random effect is assessed by comparing nested mixed effect models using likelihood ratio tests. For example, when genotype, location and year are all treated as random (Model 1), the significance of the year effect can be evaluated by fitting the following auxiliary model:
\begin{equation}
    \label{equ:aux_model}
    \begin{aligned}
    \texttt{Trait} \sim\;&
    1
    +(1 \mid \texttt{genotype})
    +(1 \mid \texttt{location}) \\
    &+(1 \mid \texttt{year}{:}\texttt{genotype})
    +(1 \mid \texttt{year}{:}\texttt{location}) \\
    &+(1 \mid \texttt{genotype}{:}\texttt{location})
    +(1 \mid \texttt{year}{:}\texttt{location}{:}\texttt{genotype}) \\
    &+(1 \mid \texttt{year}{:}\texttt{location}{:}\texttt{replication}).
    \end{aligned}
\end{equation}
Compared with Model~1, Equation~\ref{equ:aux_model} excludes only the random effect of \texttt{year}. The significance of the year effect is therefore assessed by comparing the auxiliary model with the full model. The same procedure is applied to each remaining random effect by removing the corresponding variance component while keeping all other model terms unchanged.

The two models are compared by the likelihood ratio test. Let $L_0$ and $L_1$ denote the maximum likelihoods of the auxiliary and full models. The likelihood ratio statistic is defined as
\begin{equation}
\Lambda=-2\log\left(\frac{L_0}{L_1}\right) \sim \chi^2(k), 
\end{equation}
where $k$ is the difference in the number of estimated parameters between the auxiliary and full models. According to Wilks' theorem, the likelihood ratio statistic asymptotically obeys the $\chi^2$ distribution under the null hypothesis. It should be noted that testing random effect terms involves testing whether the corresponding variance component equals zero, which is a boundary problem because variance parameters are constrained to be non-negative. The asymptotic distribution of the likelihood ratio statistic is theoretically a mixture of chi-square distributions rather than a standard one in general \cite{variance_mixed,lrt_non_standard}. In this work, we follow the common practice in linear mixed models; then the \texttt{anova} function in the \texttt{stats} package \cite{stats} is used to compute the corresponding $\chi^2$ $p$-value of each random effect.

Significance analysis identifies the fixed and random effects that contribute significantly to phenotypic variation and quantifies their corresponding variance components. In particular, the significance of GxE interactions provides the statistical justification for conducting subsequent stability analyses. Furthermore, the estimated variance components characterize the relative contributions of genotypes, environments, and their interactions to trait variation, thereby providing valuable information for variety evaluation and selection.
\subsection{Stability Analysis of Genotype-Environment Interaction}
After establishing the statistical significance of G$\times$E interactions, the next objective is to evaluate the adaptability and stability of varieties across diverse environments. This section first introduces the single-genotype stability model based on joint regression analysis, followed by the multi-genotype stability model. The single-genotype approach evaluates the stability of one genotype at a time by quantifying its response to environmental variation, whereas the multi-genotype models simultaneously investigate the adaptation patterns of all genotypes across all environments using multivariate statistical methods, including principal component analysis and singular value decomposition.
\subsubsection{Single-genotype Stability Analysis}
In the single-genotype stability model, the average trait observation (ENVTrait) needs to be calculated by grouping year-location-replication. We can assume that there are ($E_{1}, E_{2},\cdots, E_{k}$) levels of ENV and (\(R_{1}, R_{2},\cdots, R_{M}\)) levels of replication; then the following joint linear regression model is constructed: 
\begin{equation} 
{Trait}_{j} = \beta_{0j} + \beta_{1j} \cdot \text{ENVTrait} + \sum_{i = 2}^{K}\beta_{E_{ij}} \cdot I\left( \text{ENV} = E_{i} \right) + \sum_{i = 2}^{M}\beta_{R_{ij}} \cdot I\left( \text{RP} = R_{i} \right) + \epsilon_{j}, 
\end{equation} 
where $j$ represents the $j$-th genotype, \(I( \bullet )\) is an indicator function and \(\epsilon_{j}\) an error term. We take the first level \(E_{1},R_{1}\) as a reference by default.

The primary stability parameters are the regression coefficient ($\beta_{1j}$) and the deviation from regression ($s_{dj}^{2}$). The regression coefficient measures the responsiveness of genotype $j$ to changes in environmental conditions represented by the environmental index. A genotype is considered to possess broad adaptation when $\beta_{1j}$ is not significantly different from one and $s_{dj}^{2}$ is close to zero. Genotypes with $\beta_{1j}>1$ exhibit greater responsiveness to favorable environments and generally achieve superior performance under high-yield conditions, whereas genotypes with $\beta_{1j}<1$ are less sensitive to environmental variation and tend to perform more consistently under unfavorable or stress environments. Likewise, a smaller deviation from regression indicates greater phenotypic stability across environments. Accordingly, the hypotheses
\[
H_{0}:\beta_{1j}=1
\]
and
\[
H_{0}:s_{dj}^{2}=0
\]
are evaluated using the $t$-test and $F$-test, respectively. These statistical analyses are implemented using the \texttt{lm} function together with the \texttt{dplyr} \cite{dplyr} and \texttt{tidyr} \cite{tidyr} packages in R.

To compute additional stability statistics, a Gaussian linear model is fitted to estimate the residual mean square required by several classical stability measures:
\begin{equation}
    \begin{aligned}
    \texttt{Trait} \sim\;&
    \texttt{year}
    +\texttt{genotype}
    +\texttt{location} \\
    &+\texttt{year}*\texttt{location}
    +\texttt{year}*\texttt{genotype} \\
    &+\texttt{genotype}*\texttt{location} \\
    &+\texttt{year}*\texttt{location}*\texttt{genotype} \\
    &+\texttt{year}*\texttt{location}*\texttt{replication}.
    \end{aligned}
\end{equation}
On account of the estimated mean square error and genotype means, the function \texttt{stability.par} of the \texttt{agricolae} package \cite{agricolae} is employed to calculate Shukla's stability variance ($\sigma_i^{2}$), Wricke's ecovalence ($W_i^{2}$), and Kang's yield stability statistic. These complementary stability measures provide quantitative assessments of genotype stability and facilitate the identification of genotypes that simultaneously exhibit stable and high yield across multiple environments.

Although the single-genotype model provides an effective quantitative assessment of the stability of individual genotypes, it cannot simultaneously characterize the interaction patterns among all genotypes and environments. Consequently, multivariate stability models are required to investigate the overall structure of GxE interaction and to identify genotypes with superior adaptation to specific or broad environments.
\subsubsection{Multi-genotype Stability Analysis}
This section introduces the multi-genotype stability analysis models and analyzes the adaptability and capacity of multiple genotypes in multiple environments, so as to facilitate the cross-sectional comparison of multiple genotypes.

\textbf{Additive Main Effects and Multiplicative Interaction.}
AMMI integrates the strengths of analysis of variance (ANOVA) and PCA: ANOVA is first utilized to estimate the additive main effects of genotypes and environments, whereas PCA is subsequently applied to the residual G$\times$E interaction matrix to capture the underlying interaction structure through a set of multiplicative components. Consequently, AMMI partitions the observed phenotypic response into additive genotype and environment effects together with multiplicative interaction effects.

Specifically, ANOVA decomposes the phenotypic response into the grand mean, genotype main effects, and environment main effects. Then, the remaining GxE interaction residuals are subjected to SVD, yielding a sequence of interaction principal components (IPCs). The retained IPCs describe the dominant interaction patterns between genotypes and environments, while the remaining components are treated as residual variation. Suppose that the interaction matrix has rank
\[
r \leq \min(J-1, I-1),
\]
where $J$ and $I$ denote the numbers of environments and genotypes. Under a completely randomized design, the AMMI model is given by
\begin{equation}
    y_{ijk}
    =
    \mu
    +
    \alpha_i
    +
    \beta_j
    +
    \sum_{n=1}^{N}
    \lambda_n
    \gamma_{in}
    \delta_{jn}
    +
    \rho_{ij}
    +
    \varepsilon_{ijk},
\end{equation}
where $y_{ijk}$ denotes the observed trait value of genotype $i$ in environment $j$ and replication $k$; $\mu$ is the overall mean; $\alpha_i$ and $\beta_j$ represent the genotype and environment main effects, respectively; $\lambda_n$ is the singular value corresponding to the $n$-th IPC; $\delta_{jn}$ and $\gamma_{in}$ are the environment and genotype scores for $n$-th IPC; $\rho_{ij}$ denotes the residual interaction not explained by the retained IPCs; $\varepsilon_{ijk}$ is the random error; and $N \leq r$ is the number of retained interaction principal components.

AMMI can also be expressed in matrix form as
\begin{equation}
\label{equ:ammi}
    \mathbf{Y}
    =
    \mathbf{1}_I\mathbf{1}_J^{T}\mu
    +
    \boldsymbol{\alpha}\mathbf{1}_J^{T}
    +
    \mathbf{1}_I\boldsymbol{\beta}^{T}
    +
    \mathbf{UDV}^{T}
    +
    \boldsymbol{\rho}
    +
    \boldsymbol{\varepsilon},
\end{equation}
where $\mathbf{Y}\in\mathbb{R}^{I\times J}$ is the matrix of genotype means across environments. The first three terms represent the grand mean, genotype main effects, and environment main effects, respectively. After removing these additive effects, the interaction matrix
\[
\mathbf{Y}^{*}
=
\mathbf{Y}
-
\mathbf{1}_I\mathbf{1}_J^{T}\mu
-
\boldsymbol{\alpha}\mathbf{1}_J^{T}
-
\mathbf{1}_I\boldsymbol{\beta}^{T}
-
\boldsymbol{\rho}
\]
is decomposed by singular value decomposition,
\[
\mathbf{Y}^{*}
=
\mathbf{UDV}^{T},
\]
where $\mathbf{U}\in\mathbb{R}^{I\times N}$ and $\mathbf{V}\in\mathbb{R}^{J\times N}$ comprise the left and right singular vectors corresponding to genotype and environment scores, and $\mathbf{D}$ is the diagonal matrix containing the retained singular values. The residual matrix $\boldsymbol{\varepsilon}$ contains both unexplained interaction effects and experimental errors.

The selection of the number of retained interaction principal components ($N$) is critical because it directly affects model interpretation and predictive performance \cite{protocol_ammi}. Too few components may fail to capture important interaction patterns, whereas too many components may incorporate random noise and lead to overfitting. Forkman and Piepho proposed a parametric bootstrap procedure for testing the significance of multiplicative interaction terms and determining the appropriate model rank \cite{parametric_bootstrap}. Following this approach, the GxEStat package implements the AMMI model and its associated biplot visualization using the function \texttt{AMMI} in the \texttt{agricolae} package \cite{agricolae}.

\textbf{Genotype Main Effects plus Genotype-Environment Interaction.}
Unlike the AMMI, which separately models genotype, environment, and G$\times$E interaction effects, the GGE model combines genotype main effects (G) and the G$\times$E interactions (GE) into a single multiplicative component while removing the environment main effect. The rationale behind this formulation is that, for variety evaluation and selection, the environment main effect is generally not informative because it is common to all genotypes, whereas genotype performance is determined jointly by genotype main effects and GxE interactions. Consequently, GGE biplots simultaneously display genotype performance and stability, providing an effective graphical tool for variety evaluation and mega-environment identification.

GGE biplots are constructed by applying SVD to the environment-standardized genotype $\times$ environment data matrix. In practice, only the first two principal components are retained because they usually explain the majority of the GGE variation. The first principal component (PC1) primarily reflects the average performance (or fitness) of genotypes across environments, whereas the second principal component (PC2) mainly characterizes genotype stability and crossover GxE interactions.

The GGE model is formulated as
\begin{equation}
    \mathbf{Y}
    =
    \mathbf{X}_{1}\boldsymbol{\beta}
    +
    \sum_{k=1}^{t}
    \lambda_{k}
    \operatorname{diag}
    \left(
    \mathbf{Z}\boldsymbol{\alpha}_{k}
    \right)
    \mathbf{X}_{2}
    \boldsymbol{\gamma}_{k}
    +
    \boldsymbol{\varepsilon},
\end{equation}
where $\mathbf{Y}$ denotes phenotypic observation data with dimension $v\times(r\times j)$, in which $v$, $r$ and $j$ are the numbers of genotypes, replications and environments, respectively. The vector $\boldsymbol{\beta}$ contains the fixed effects of replications within environments estimated from the corresponding mixed-effects model. For the $k$-th principal component, $\lambda_k$ denotes the singular value, whereas $\boldsymbol{\alpha}_k$ and $\boldsymbol{\gamma}_k$ represent the genotype and environment singular vectors. The number of retained principal components is
\[
t=\min(v-1,j),
\]
which corresponds to the rank of the GGE matrix. Furthermore, $\mathbf{X}_1$, $\mathbf{X}_2$, and $\mathbf{Z}$ are the associated design matrices.
The residual vector satisfies
\[
    \boldsymbol{\varepsilon}
    \sim
    N\!\left(
    \mathbf{0},
    \sigma_e^2\mathbf{I}
    \right),
\]
where $\sigma_e^2$ is the residual variance and $\mathbf{I}$ is the identity matrix. Consequently, the response vector follows
\begin{equation}
    \mathbf{Y}
    \mid
    \boldsymbol{\beta},
    \lambda,
    \boldsymbol{\alpha},
    \boldsymbol{\gamma},
    \sigma_e^2
    \sim
    N
    \left(
    \boldsymbol{\mu},
    \sigma_e^2\mathbf{I}
    \right),
\end{equation}
with
\[
    \boldsymbol{\mu}
    =
    \mathbf{X}_{1}\boldsymbol{\beta}
    +
    \sum_{k=1}^{t}
    \lambda_{k}
    \operatorname{diag}
    \left(
    \mathbf{Z}\boldsymbol{\alpha}_{k}
    \right)
    \mathbf{X}_{2}
    \boldsymbol{\gamma}_{k}.
\]

Compared with the AMMI biplot, GGE biplots provide a more direct graphical method for genotype evaluation because both genotype main effects and G$\times$E interactions are jointly represented. Based on the geometric relationships among genotype and environment vectors, the GGE biplot supports a variety of graphical analyses, including ``which-won-where'' patterns, mean performance versus stability, discriminating ability and representativeness of environments, and genotype ranking relative to an ideal genotype. These visualizations facilitate the identification of broadly adapted genotypes, specifically adapted genotypes, and potential mega-environments. In GxEStat, the GGE model is implemented using the \texttt{GGEBiplotGUI} package \cite{interactive_biplot}.

\section{Experiment Results and Analysis}
\begin{figure}[t]
    \centering
    \includegraphics[width=1.0\linewidth]{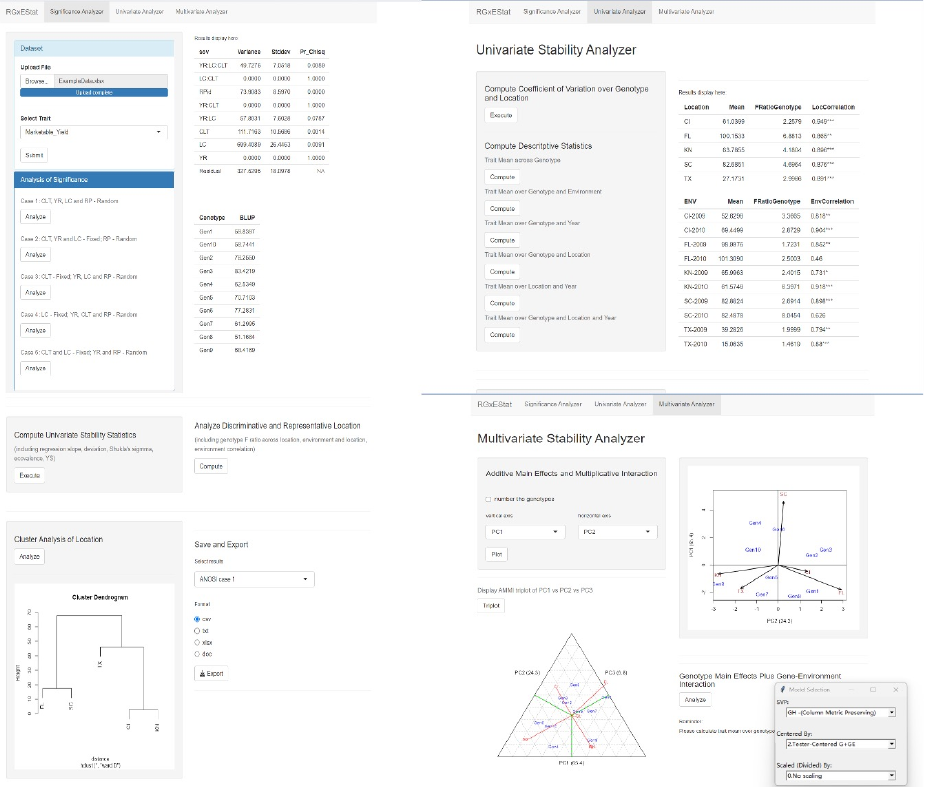}
    \caption{Interface of our GxEStat.}
    \label{fig:interface}
\end{figure}

Here, we will use GxEStat to perform the significance and stability analysis on two publicly available breeding datasets and show the experimental results of the models. The interface of GxEStat is shown in Figure \ref{fig:interface}, including significance analysis, single-genotype stability analysis, and multi-genotype stability analysis.
\subsection{Data Sources}
\begin{table}[t]
    \centering
    \small
    \caption{Demo of watermelon yield trial data in the southern United States in 2009--2010.}
    \label{tab:watermelon_demo}
        \begin{tabular}{ccclr}
        \toprule
        year & location & replication & genotype & marketable yield \\
        \midrule
        2009 & KN & 1 & EarlyCanada        & 56.236 \\
        2009 & KN & 1 & CalhounGray       & 74.167 \\
        2009 & KN & 1 & GeorgiaRattlesnake& 55.873 \\
        2010 & TN & 2 & EarlyCanada        & 32.601 \\
        2010 & TN & 2 & CalhounGray       & 74.167 \\
        2010 & TN & 2 & GeorgiaRattlesnake& 64.794 \\
        \bottomrule
        \end{tabular}
\end{table}
The experiment data selected in this paper consists of two parts: watermelon breeding data in the southern part of the US from 2009 to 2010 \cite{watermelon_breeding_data}; oat field random trial data from agridat \cite{agridat}. The watermelon breeding data in the southern United States were obtained from random and replicated block trials across 2 years, 10 varieties, 5 locations, 4 replications, with the marketable yield as the analyzed trait. The oat field trial data includes 6 environments, 24 varieties and the corresponding marketable yield. 
Table \ref{tab:watermelon_demo} shows the template of multi-year multi-location random trial data (breeding data) of genotype and environment.
\subsection{Significance Analysis Results}
This section begins with modeling watermelon breeding data by constructing the mixed effect model of Model 1. Subsequently, we perform a significant analysis of all the fixed and random terms in the model by the F test and likelihood ratio test respectively, and determine the significant effects influencing the watermelon yield. The analysis results are shown in Table \ref{tab:variance_component}.
\begin{table}[t]
    \centering
    \small
    \caption{Significance analysis of watermelon breeding data.}
    \label{tab:variance_component}
    \begin{tabular}{lccc}
    \toprule
    Source of variation &
    Variance &
    Standard deviation &
    \makecell{Likelihood ratio $p$-value} \\
    \midrule
    year $\times$ location $\times$ genotype & 49.74  & 7.05  & 0.008 \\
    year $\times$ location $\times$ replication  & 73.91  & 8.60  & 0.000 \\
    location $\times$ genotype             & 0.00   & 0.00  & 1.000 \\
    year $\times$ genotype             & 0.00   & 0.00  & 1.000 \\
    year $\times$ location              & 57.81  & 7.60  & 0.078 \\
    genotype                         & 111.68 & 10.57 & 0.001 \\
    location                          & 699.36 & 26.45 & 0.009 \\
    year                          & 0.00   & 0.00  & 1.000 \\
    residual                    & 327.53 & 18.10 & -- \\
    \bottomrule
    \end{tabular}
\end{table}
In this example, the likelihood ratio tests indicate that the random effects corresponding to year, location $\times$ genotype, year $\times$ genotype, and year $\times$ location are not statistically significant ($p>\alpha=0.05$). Hence, there is insufficient statistical evidence that these variance components contribute to the variability in watermelon yield. The results further suggest that the proposed random-effect structure may be unnecessarily complex relative to the available data. In practice, variance components estimated at or near zero frequently occur when the corresponding random factors have only a few levels or provide limited information for variance estimation. Under such circumstances, Bayesian mixed-effect modeling using the Markov chain Monte Carlo method (MCMC), as computed in the MCMCglmm package \cite{mcmc}, offers a robust alternative for estimating variance components and quantifying the uncertainty associated with random effects through their posterior distributions.

Subsequently, the oat field trial data were analyzed using the mixed-effects model described in Model 2, and the results are presented in Table~\ref{tab:oat_significance}. Since all observations were collected from a single year, no year-related effects were incorporated into the model. The F-tests show that genotype (F=3.547, p=0.013), location (F=2.856, p=0.037), and the genotype $\times$ location interaction (F=2.989, p=0.018) are all statistically significant at the 5\% significance level. Therefore, both genotype and location contribute significantly to the observed variation in oat yield. Moreover, the significant genotype $\times$ location indicates that the relative performance of genotypes depends on the testing environment, highlighting the importance of accounting for G×E interaction in varietal evaluation and selection.
\begin{table}[t]
    \centering
    \small
    \caption{Significance analysis of oat field data.}
    \label{tab:oat_significance}
    \begin{tabular}{lccc}
    \toprule
    Source of variation &
    Mean square &
    $F$ statistic &
    $p$-value \\
    \midrule
    location $\times$ genotype & 0.411 & 2.989 & 0.018 \\
    genotype             & 0.554 & 3.547 & 0.013 \\
    location              & 0.334 & 2.856 & 0.037 \\
    residual        & 0.156 & --    & --    \\
    \bottomrule
    \end{tabular}
\end{table}
\begin{figure}[t]
    \centering
    \includegraphics[width=1.0\linewidth]{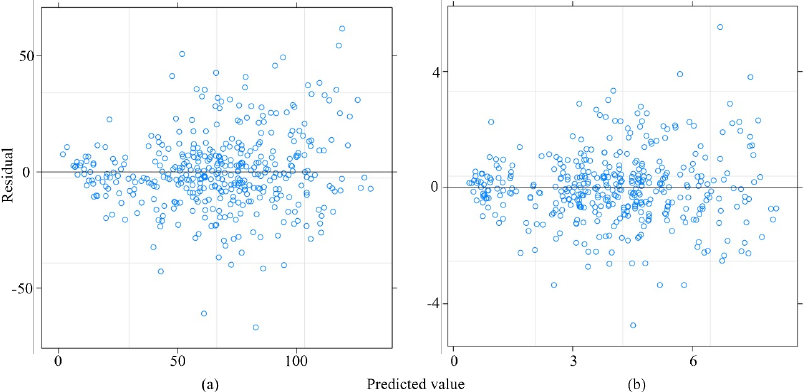}
    \caption{Scatter plots of yield predictions versus model residuals for (a) watermelon breeding data and (b) oat field random trial data.}
    \label{fig:scatter_plot}
\end{figure}

Finally, the GxEStat can automatically calculate the BLUPs of fixed and random effects in the model and get the yield estimates. The predictions of yield and the residual error of the model are shown in Figure \ref{fig:scatter_plot}. After significant analysis, we can judge that genotype and genotype-environment effects will affect the yield of watermelon and oat, so it is necessary to implement subsequent stability analysis to select varieties with excellent high yield and strong stability.
\subsection{Results of Single-genotype Stability Analysis}
In this section, we will show GxEStat to model the single-genotype stability of watermelon breeding data. GxEStat can fast obtain single-genotype stability statistics, including single-genotype regression slope (\(\beta_{1j}\)), regression deviation (\(s_{d}^{2}\)), Shukla's \(\sigma_{i}^{2}\), ssquares, Wricke's stability ecovalence (\(W_{i}^{2}\)) and Kang's yield stability statistic (\(YS_{i}\)). The results of single-genotype stability analysis of watermelon breeding data are shown in Table \ref{tab:watermelon_stability}.
\begin{table}[t]
    \centering
    \small
    \caption{Single-genotype stability analysis of watermelon data.}
    \label{tab:watermelon_stability}
    \begin{tabular}{lcccccc}
    \toprule
    genotype &
    $\beta_{1i}$ &
    $s_{di}^{2}$ &
    $\sigma_{i}^{2}$ &
    $s_i^{2}$ &
    $W_i^{2}$ &
    $YS_i$ \\
    \midrule
    CalhounGray        & 1.301   & 124.670      & 61.347$^{\mathrm{ns}}$   & 15.761$^{\mathrm{ns}}$   & 279.747  & 10+ \\
    CrimsonSweet       & 1.341   & 1450.035$^{**}$   & 439.125$^{\mathrm{ns}}$  & 567.988$^{\mathrm{ns}}$  & 1488.636 & 4   \\
    EarlyCanada        & 0.249$^{**}$ & 686.251$^{**}$    & 253.230$^{\mathrm{ns}}$  & 285.370$^{\mathrm{ns}}$  & 893.772  & 2   \\
    FiestaF1           & 1.639   & 657.873      & 300.285$^{\mathrm{ns}}$  & 385.997$^{\mathrm{ns}}$  & 1044.349 & 11+ \\
    GeorgiaRattlesnake & 0.945   & 220.063      & 52.213$^{\mathrm{ns}}$   & 44.863$^{\mathrm{ns}}$   & 250.518  & 8+  \\
    Legacy             & 1.056   & 428.070      & 287.394$^{\mathrm{ns}}$  & 262.488$^{\mathrm{ns}}$  & 1003.097 & 7+  \\
    Mickylee           & 0.618   & 705.481$^{**}$    & 188.105$^{\mathrm{ns}}$  & 195.126$^{\mathrm{ns}}$  & 685.374  & 3   \\
    Quetzali           & 0.965   & 96.532       & 82.809$^{\mathrm{ns}}$   & 86.103$^{\mathrm{ns}}$   & 348.425  & 1   \\
    StarbriteF1        & 1.388   & 221.137      & 157.241$^{\mathrm{ns}}$  & 78.371$^{\mathrm{ns}}$   & 586.607  & 12+ \\
    SugarBaby          & 0.498$^{**}$ & 332.181$^{**}$    & 264.187$^{\mathrm{ns}}$  & 308.430$^{\mathrm{ns}}$  & 928.836  & -1  \\
    \bottomrule
    \end{tabular}
\end{table}

In Table \ref{tab:watermelon_stability}, ${\mathrm{ns}}$ indicates not significant, and + means that the genotype can be judged to be stable according to Kang's yield stability. In the \(\beta_{1j}\) column, $**$ indicates that the t-test p-values of the regression slope corresponding to this genotype is less than the significance level \(\alpha = 0.01\) ($***$ and $*$ are the significance levels of 0.001 and 0.05), so \({H0:\ \beta}_{1j} = 1\) is rejected, stating that the regression slope is significantly different from 1. Similarly, in \(s_{d}^{2}\) column, $**$ represents F-test p-value of the regression deviation corresponding to the genotype is less than the significant level \(\alpha = 0.01\), and the rejection of \(H0:s_{d}^{2} = 0\) indicates that the regression deviation is significantly different from 0. From the analysis in the table, it can be concluded that the watermelon varieties of CalhounGray, FiestaF1, GeorgiaRattlesnake, Legacy, and StarbriteF1 have strong adaptability to the environment and are stable genotypes.
\subsection{Results of Multi-genotype Stability Analysis}
In this section, we mainly present the principal component biplots of stability analysis model, and utilize them to compare the superiority between multiple genotypes and multiple environments, and determine the mutual assistance between genotypes and environments.
\subsubsection{Results on Additive Main Effect and Multiplicative Interaction}
\begin{figure}[t]
    \centering
    \includegraphics[width=1.0\linewidth]{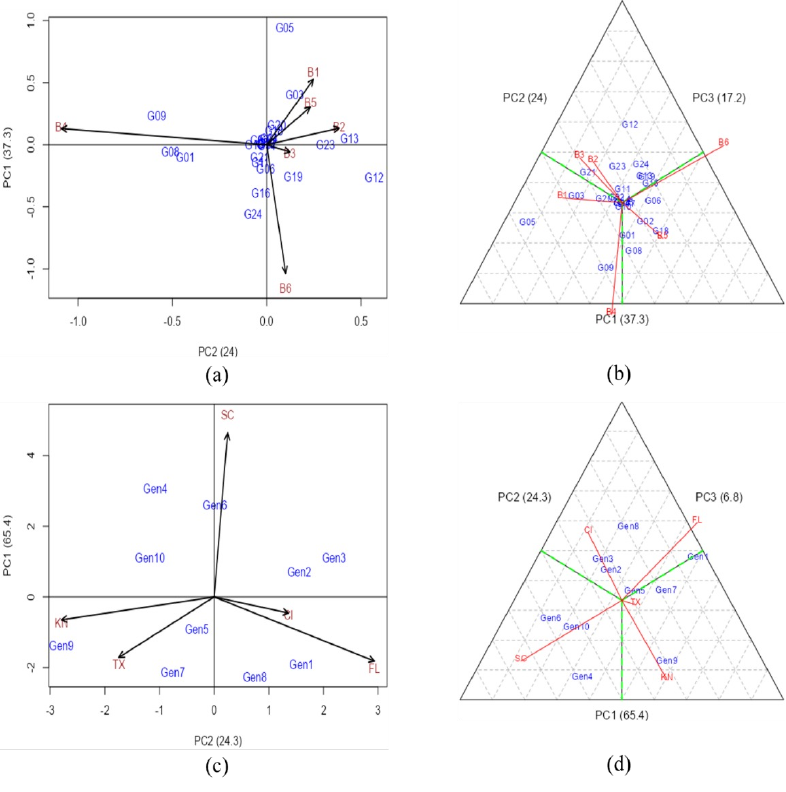}
    \caption{(a) PC1 v.s. PC2 biplot of oat field data, (b) PC1 v.s. PC2 v.s. PC3 triplot of oat field data, (c) PC1 v.s. PC2 biplot of watermelon breeding data. (d) PC1 v.s. PC2 v.s. PC3 triplot of watermelon breeding data.}
    \label{fig:ammi}
\end{figure}

Taking watermelon breeding data and oat field data as examples, the biplots of AMMI model drawn by GxEStat is shown in Figure \ref{fig:ammi}, where the number of principal components retained is set to 4.

In order to make biplots more concise, Gen1-10 is used to replace watermelon varieties EarlyCanada, CalhounGray, StarbriteF1, CrimsonSweet, GeorgiaRattlesn, FiestaF1, Mickylee, SugarBaby, Legacy and Quetzali respectively in this section and thereafter. According to Equation~\ref{equ:ammi}, the interactions between genotypes and environments can be represented by multiplying the genotype score and environment score in each principal component and summing them along the principal components. Figure \ref{fig:ammi} provides a very effective depiction of the genotype-environment interactions. In Figure \ref{fig:ammi}, genotypes and environments near the axis origin will not interact with each other (the product of their scores will be close to zero); genotypes and environments far away from the axis origin show great interaction, and therefore have high yield instability. Some people argue that the Euclidean distance from the origin should be regarded as a measure of instability. When genotypes and environments are close to one another, the interaction is positive. If two objects are close, their scores (coordinates) will have the same sign, so their products will be positive. When genotypes and environments are far away from each other, the interaction is negative.

For example, for oat varieties, G03 variety in B1 environment, G13 in B2 environment, G05 in B1, B5, B2 environment have good yields (compared with the average level), while watermelon varieties Gen7 and Gen9 have particularly excellent yields in KN and TX environment.

\subsubsection{Results on Genotype Main Effect plus Genotype-Environment Interaction}
\begin{figure}[htbp]
    \centering
    \includegraphics[width=0.9\linewidth]{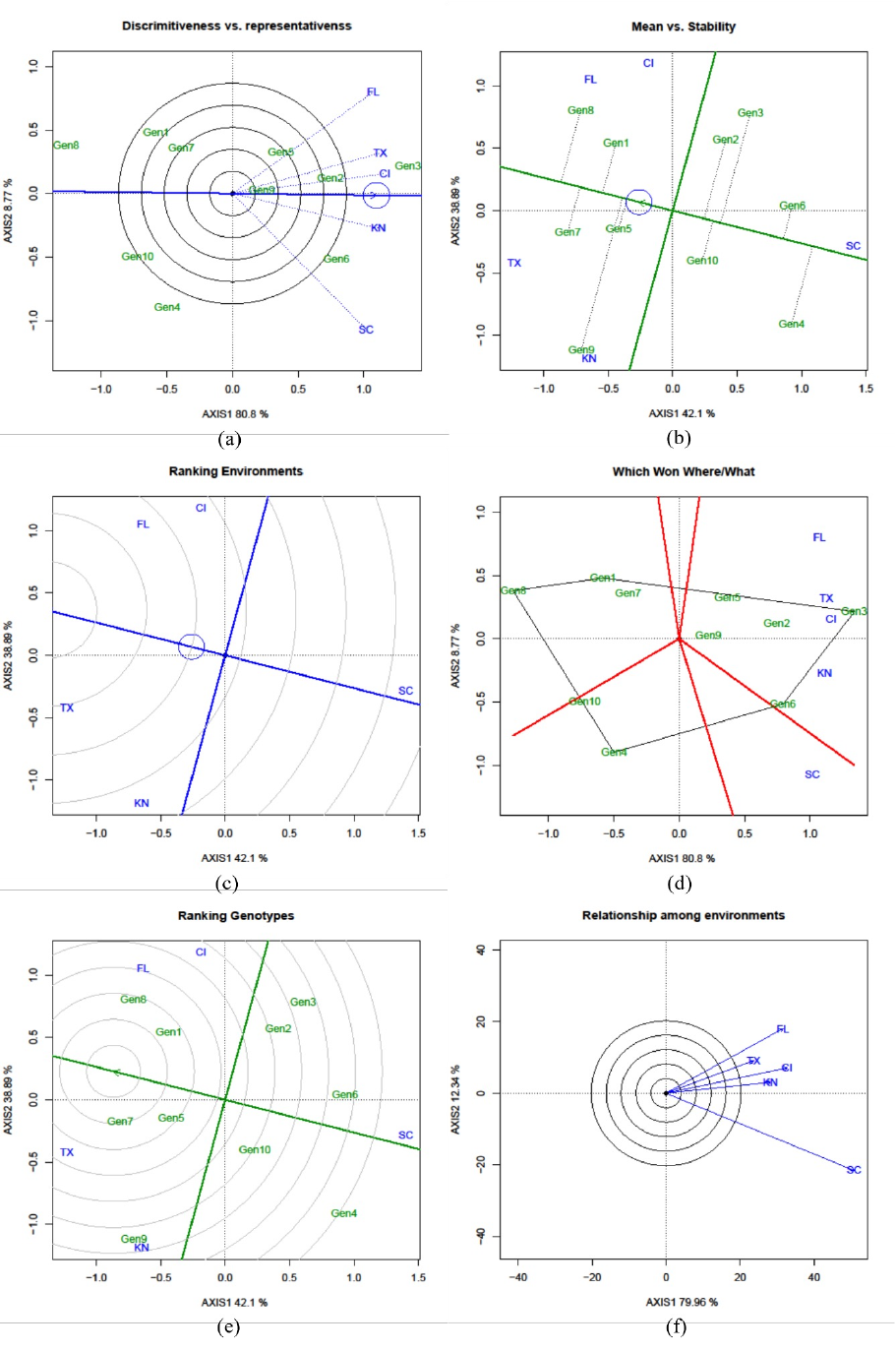}
    \caption{GGE biplots: (a) discrimitiveness v.s. representativeness, (b) yield mean v.s. stability, (c) ranking environments, (d) which won where/what, (e) ranking genotypes, (f) relationship among environments.}
    \label{fig:gge}
\end{figure}

The GGE decomposes the total effects of genotypes and genotype-environment interactions by SVD and PCA, and then makes various biplots according to PC1 and PC2. In this section, only the watermelon breeding data is taken as an example to show the analysis process of GGE model by GxEStat. GxEStat can plot various GGE biplots with one click to identify varieties with high stability and high yield, and can also suggest which varieties are suitable for planting in a specific environment. Figure \ref{fig:gge} illustrates a wide variety of GGE biplots, including discrimitiveness and representativeness, variety yield and stability, environment ranking, which won where/what, variety ranking and relationship between environments. The PC1 v.s. PC2 biplot is not shown here, because it is similar to Figure \ref{fig:ammi} (c).

Figures \ref{fig:gge} (a) and (c) are used to judge the discrimination and representativeness of different environments, i.e., which environment can better distinguish high-yield and high-stability varieties (the length of line segment in Figure \ref{fig:gge} (a)) and which environment has strong representativeness for the target ecological zone. The arrowed line in (a) is the mean environment axis. The direction indicated by the arrowhead on it is the evaluation of the discrimination and representativeness of the test locations. The angle between the line segment of the test site and the mean environmental axis is a measure of its representativeness of the target environment. The smaller the angle, the stronger the representativeness. If the angle between a test site and the mean environment axis is obtuse, it is not suitable as a trial environment. A test site without discrimination is useless. Discriminatory but unrepresentative test sites can eliminate unstable varieties. Only those with both discriminative and representative test sites can choose the best varieties with high and stable yield, such as the environment CL and TX in the biplot. Figure \ref{fig:gge} (c) draws concentric circles according to the mean environment axis, where the smaller the circle is, the stronger the distinctiveness and representativeness of the environment.

Figure \ref{fig:gge} (b) and (e) reflect the productivity and stability of varieties. In Figure \ref{fig:gge} (e), concentric circles are made according to the mean environment axis and if the circle where varieties are located is smaller, the better yield-environment interaction of varieties is, the better the variety is. Similarly, Figure \ref{fig:gge} (b), mean v.s. stability, also requires the mean environment axis (straight line with arrow) and the mean environment value. There is also a straight line perpendicular to the mean environment axis through the center. The direction of the mean environment axis refers to the trend of the approximate average yield of varieties in all environments. Make a vertical line (green dotted line) between varieties and the mean environment axis. The longer the vertical line between varieties and the mean environment axis, the more unstable varieties are. Thus, Gen1, Gen5 and Gen7 have stability and high yield.

Figure \ref{fig:gge} (d) mainly illustrates the variety with the highest yield in each environment according to the interaction between the genotype and the environment. The figure connects the farthest points in all directions with straight lines to form a polygon, and partitions the biplot into several sections by making the center perpendicular to each side. The variety is distributed in the sectors. For the environments in a specific section, the genotype at the top corner of the section has the best yield. For example, Gen3 variety has the highest yield in FL, TX, CL, and KN environments.

Figure \ref{fig:gge} (f) depicts the relationship between the environments. It draws a line segment from the coordinate center to each environment, mainly to evaluate the distinctiveness and similarity of the environments. The cosine value of the angle (\(0^{{^\circ}} - 180^{{^\circ}}\)) between the two environment line segments is their correlation coefficient. The angle between the two line segments is less than $90^{{^\circ}}$, which shows a positive correlation, and the two environments are similar in ranking the variety quality. The angle between the two line segments greater than 90 degrees indicates negative correlation, and the ranking of variety quality in the two environments is mostly opposite; The angle equal to 90 degrees means that the two environments are not related. The smaller angle close to zero degrees means that the two environments may be duplicated trial sites, and the evaluation of varieties will not be affected if one of them is removed. The length of the line segment stands for the ability of the environment to discriminate between varieties; the longer the line segment is, the stronger the discrimination is.

\section{Conclusion and Discussion}
This paper presents GxEStat, a unified computational statistical framework for G$\times$E interaction analysis. The proposed framework integrates significance testing based on linear mixed effect models, single-genotype and multi-genotype stability models into a reproducible and user-friendly analytical workflow. GxEStat reduces the technical barriers associated with conventional command-line statistical software and facilitates efficient analysis of multi-environment trial data. We expect that it can serve not only as a practical analysis platform but also as a benchmark environment for evaluating emerging statistical methodologies for G$\times$E interaction analysis.

Several research directions deserve further investigation. First, current G$\times$E analyses are primarily based on linear or mixed-effects statistical models, which may have limited capability in capturing complex nonlinear interactions among genotypes, environments, and management practices. Future work may integrate modern statistical learning and machine learning approaches with classical mixed-effects modeling to improve predictive performance while preserving statistical interpretability. Second, most existing methods focus on the analysis of a single phenotypic trait. In practical breeding programs, however, breeders simultaneously optimize multiple correlated traits measured across multiple environments. Developing statistically rigorous computational methods for joint modeling of multi-trait, multi-environment, and potentially high-dimensional genomic data remains an important open problem. Finally, with the increasing genomic, phenomic, environmental, and remote-sensing data, future statistical software should support scalable algorithms capable of integrating heterogeneous data sources, thereby enabling comprehensive analyses for next-generation precision breeding.

\section*{Acknowledgements}
The authors gratefully acknowledge Wehner et al. from North Carolina State University, for making the watermelon breeding datasets publicly available. We also thank the developers of the agridat package for compiling and maintaining the publicly available oat breeding dataset used in this study. Finally, we sincerely thank Prof. Yanfeng Sun from the State Key Laboratory of Cotton Biology, Henan University, for the constructive suggestions.









\bibliography{ref}

\end{document}